\documentclass[10pt,twocolumn,letterpaper]{article}

\usepackage{wacv}              
\usepackage{graphicx}
\usepackage{amsmath}
\usepackage{amssymb}
\usepackage{booktabs}
\usepackage{multirow}
\usepackage{multicol}
\usepackage{makecell}
\usepackage[accsupp]{axessibility}
\usepackage{easyReview}
\usepackage[pagebackref,breaklinks,colorlinks]{hyperref}

\setreviewsoff
\usepackage[capitalize]{cleveref}
\crefname{section}{Sec.}{Secs.}
\Crefname{section}{Section}{Sections}
\Crefname{table}{Table}{Tables}
\crefname{table}{Tab.}{Tabs.}

\def\wacvPaperID{725} 
\def\confName{WACV}
\def\confYear{2024}

\begin{document}

\title{SAM Fewshot Finetuning for Anatomical Segmentation in Medical Images}

\author{
  Weiyi Xie, Nathalie Willems, Shubham Patil, Yang Li, and Mayank Kumar\\
  Stryker AI Research\\
  {\tt\small \{weiyi.xie,nathalie.willems,shubham.patil,yang.li1,mayank.kumar\}@stryker.com}
}
\maketitle

\begin{abstract}

We propose a straightforward yet highly effective few-shot fine-tuning strategy for adapting the Segment Anything (SAM) to anatomical segmentation tasks in medical images. Our novel approach revolves around reformulating the mask decoder within SAM, leveraging few-shot embeddings derived from a limited set of labeled images (few-shot collection) as prompts for querying anatomical objects captured in image embeddings. This innovative reformulation greatly reduces the need for time-consuming online user interactions for labeling volumetric images, such as exhaustively marking points and bounding boxes to provide prompts slice by slice. With our method, users can manually segment a few 2D slices offline, and the embeddings of these annotated image regions serve as effective prompts for online segmentation tasks. Our method prioritizes the efficiency of the fine-tuning process by exclusively training the mask decoder through caching mechanisms while keeping the image encoder frozen. \add{Importantly, this approach is not limited to volumetric medical images, but can generically be applied to any 2D/3D segmentation task}. 

To thoroughly evaluate our method, we conducted extensive validation on four datasets, covering six anatomical segmentation tasks across two modalities. Furthermore, we conducted a comparative analysis of different prompting options within SAM and the fully-supervised nnU-Net. The results demonstrate the superior performance of our method compared to SAM employing only point prompts ($\sim$50\% improvement in IoU) and performs on-par with fully supervised methods whilst reducing the requirement of labeled data by at least an order of magnitude. 

\end{abstract}

\section{Introduction}
\label{sec:intro}
Medical imaging such as CT, MRI, and X-Ray, etc are the most effective technique for in vivo analysis of diverse human anatomical structures. However, visual assessment of anatomical structures, even by experts, can introduce subjectivity, errors, and significant delays. Therefore, there is growing interest in leveraging computational approaches to automatically analyze medical images. In this regard, automated anatomical segmentation methods have become essential, enabling precise identification and delineation of regions of interest (ROI) before deriving clinical measures. \par

In recent years, several automatic segmentation methods \cite{ronneberger2015u,cicek20163d,milletari2016v,chen2018drinet,isensee2021nnu,cao2022swin,hatamizadeh2022unetr} have made significant strides in addressing the limitations associated with manual segmentation, resulting in improved reliability, reproducibility, and efficiency in the analysis of medical images \cite{isensee2021nnu}. These approaches are based on the latest advances in deep learning with diverse architectures mostly following the design of UNet (\cite{isensee2021nnu}) and transformers \cite{vaswani2017attention,dosovitskiy2020image}). \par
The current approach to segmenting anatomical structures in medical images relies on task-specific neural networks tailored to predefined anatomical targets. However, these models struggle to generalize when encountering unfamiliar or diseased anatomies. Consequently, practitioners often face the need to develop new models with a new round of data collection and labeling, which is particularly expensive for large volumetric medical datasets such as CT and MRI. Hence, leveraging pre-trained segmentation foundation models for segmenting medical images is a fundamental research question. These pre-trained foundation models should be rich in model capacity to represent complex anatomical structures, having been trained on diverse and extensive datasets.

Indeed, segmentation foundation models have become a valuable framework for transfer learning and domain adaptation \cite{kirillov2023segment,zou2023segment,zou2023generalized}, demonstrating outstanding segmentation performance on major natural image benchmarks \cite{zou2023segment}. Compared to smaller models trained on limited data, these models exhibit comprehensive representation capabilities, suggesting their potential for improved generalization on new tasks. Segment Anything (SAM) model \cite{kirillov2023segment}, one of the prominent segmentation foundation models, has gained widespread recognition. Using an extensive dataset comprising over one billion masks, SAM demonstrates impressive zero-shot proficiency in generating precise object masks for unseen tasks. 

\begin{figure}[t]
\centering
\includegraphics[width=1.0\linewidth]{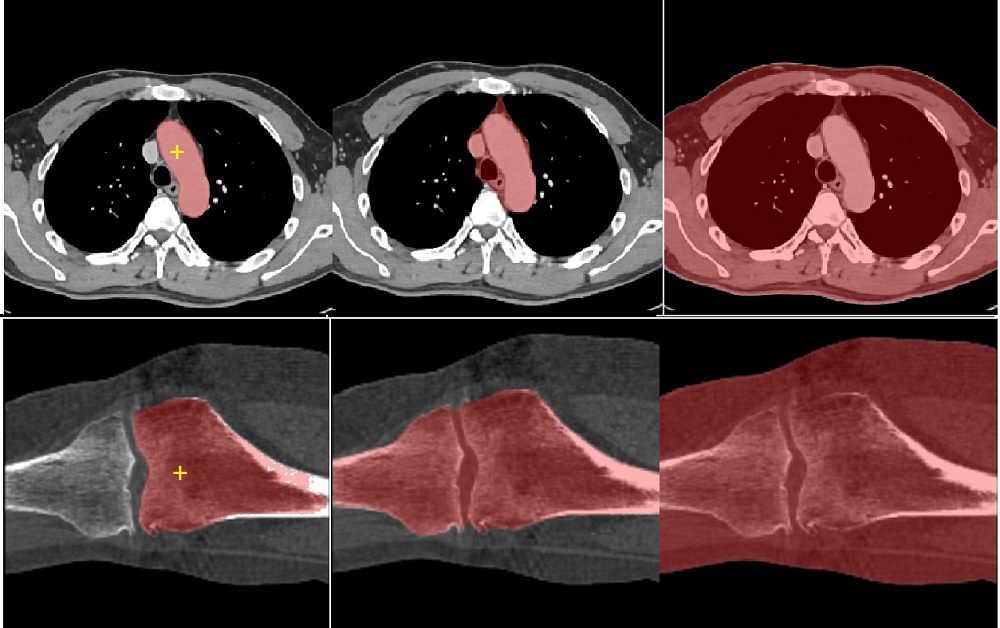}
   \caption{Two cases using SAM point prompting. The points are marked in yellow. Segmentation is colored red. The top row is an axial slice from a chest CT scan with a segmented aorta, and the bottom row is a coronal slice of a CT image with a segmented femur. We show three segmentation predictions from SAM for each case, with predicted IoU and stability score above the default thresholds (0.88 and 0.95, respectively).
  }
   \label{fig:SAM_point_fail}
\end{figure}

SAM provides versatile options for prompting: bounding boxes, points, masks, or texts. With these prompting methods, SAM promises to reduce the need for task-specific data to fine-tune and retrain new models when applied to a novel task. However, prompting may pose challenges when adapting SAM for segmenting volumetric medical images such as CT or MRI. First, SAM is inherently a 2D model. For segmenting large 3D volumetric scans, annotating regions by adding points or boxes slice by slice is still time consuming. Our experiments demonstrate that point-only prompting yields subpar performance when segmenting typical anatomies in CT and MRI images. Achieving good zero-shot performance requires accurate bounding boxes on each object region or sub-regions, substantially increasing the prompting efforts. Prompting is even more challenging when anatomical structures exhibit considerable shape variation or are distributed in multiple disconnected areas in 2D cross-section views. Second, SAM's ability to segment anything inherently leads to ambiguous predictions, particularly when anatomical structures appear closely layered in 2D views. As illustrated in Figure~\ref{fig:SAM_point_fail}, relying solely on a single point prompt is insufficient for accurately segmenting the aorta (top) and knee joints (bottom) in CT images. Even with the inclusion of bounding boxes as prompts, accurately distinguishing the aorta from surrounding arteries remains challenging because the box would cover all anatomies nearby. In this case, one has to provide multiple points and distinguish these as foreground or background to exclude surrounding objects. This increases the prompting efforts dramatically. Finally, in addition to providing three segmentation predictions, SAM incorporates two estimations, namely Intersection Over Union (IoU) and stability score, to assist users in evaluating the reliability of the predictions. However, these measurements may not be sufficient for users to determine which segmentation to select confidently for end-application. This limitation is evident in two examples illustrated in Figure~\ref{fig:SAM_point_fail}, where the last column depicts the segmentation predictions with the highest predicted IoUs and stability scores. In these examples, it is observed that the predictions with the highest scores tend to correspond to segments that are relatively easier to delineate, often achievable through simple intensity thresholding. Consequently, relying solely on the IoU and stability scores may lead to sub-optimal segmentation. To ensure accurate results, precise prompting and careful selection among three predictions become crucial, requiring users to participate.

This paper proposes a few-shot fine-tuning strategy for adapting SAM to segment anatomical structures in medical images. Importantly, this proposal does not involve introducing new network architectures or models. Instead, all components utilized in this study are from the original SAM. The key modification lies in reformulating SAM's mask decoder, which is adapted to accept few-shot embeddings as prompts, eliminating positional-encoded points, bounding boxes, or dense prompts such as masks and their corresponding prompt encoders. The fine-tuning process focuses on training SAM's mask decoder on a small set of labeled images specific to the segmentation task. Our fine-tuning process is computationally efficient compared to training standalone neural networks. To validate the effectiveness of our approach, we conducted a comprehensive evaluation comparing it with various prompting options offered by SAM, as well as a fully supervised nnU-Net trained on all available labeled images. Our evaluation was conducted on six anatomical structures. The findings indicate that the proposed fewshot fine-tuning method achieves anatomical structure segmentation performance comparable to SAM when using accurate bounding boxes as prompts while significantly outperforming SAM when using foreground points alone as prompts. 

\section{Related Work}\label{sec:related_work}
Since the introduction of SAM, several recent studies \cite{cheng2023sam,he2023accuracy} have investigated its performance in medical image segmentation benchmarks, specifically comparing various prompting options. Most of these studies suggest that using bounding boxes as prompts generally leads to improved performance compared to using points alone, though this finding is inconsistent depending on the dataset. In line with this ongoing discussion, our paper delves into the intricacies of employing SAM with varying modes of prompting, providing further insights and analysis.

In adapting SAM for medical image segmentation, MedSAM \cite{ma2023segment} performs fine-tuning of SAM's mask decoder using a carefully curated medical image dataset comprising over 200,000 masks from 11 different modalities. Notably, their fine-tuning process focuses solely on the mask decoder while keeping the remaining components of SAM frozen. The results demonstrate significant improvements in segmentation performance compared to zero-shot SAM with various prompting options. However, the effectiveness of MedSAM has not been validated against fully supervised methods. Additionally, the training effort required for fine-tuning, including data collection and computation time, poses practical challenges in adopting their approach.

Another approach, Med SAM Adaptor (MSA) \cite{wu2023medical}, enhances SAM by introducing a set of adaptor neural network modules connected through the original SAM network modules. During training, the SAM modules remain unchanged, while the parameters of the adaptor modules are updated to achieve the goal of segmenting medical images. MSA exhibits promising results in adapting SAM for the medical domain, demonstrating its effectiveness on 19 medical image segmentation datasets. However, the training process of MSA still entails a non-trivial cost. Additionally, relatively large dataset collection is needed for fine-tuning. In contrast, our proposed approach maintains the integrity of SAM's image encoder while focusing solely on fine-tuning the mask decoder using a minimal amount (5-20) of labeled images specific to the given segmentation task. Our approach significantly reduces the training effort required and provides a practical solution for adapting SAM to medical image segmentation.

Most of the existing approaches adapting SAM for medical images are still prompting-based methods, requiring users to provide accurate prompts during the use of the algorithms that may not be ideal for large volumetric medical images.

\section{Method}\label{sec:method}
This section first describes the SAM method briefly before introducing the proposed fewshot fine-tuning strategy. 
\subsection{Segment Anything}
The Segment Anything Model (SAM) is a state-of-the-art segmentation model trained on the largest segmentation dataset to date \cite{kirillov2023segment}. SAM was extensively evaluated on 23 diverse datasets upon its release, covering a wide range of natural images \cite{kirillov2023segment}. The evaluation demonstrated that SAM achieved remarkable accuracy in zero-shot applications, outperforming other interactive or dataset-specific models without requiring re-training or fine-tuning on unseen datasets or segmentation tasks.

\begin{figure*}[t]
  \centering
   \includegraphics[width=0.93\linewidth, height=9cm]{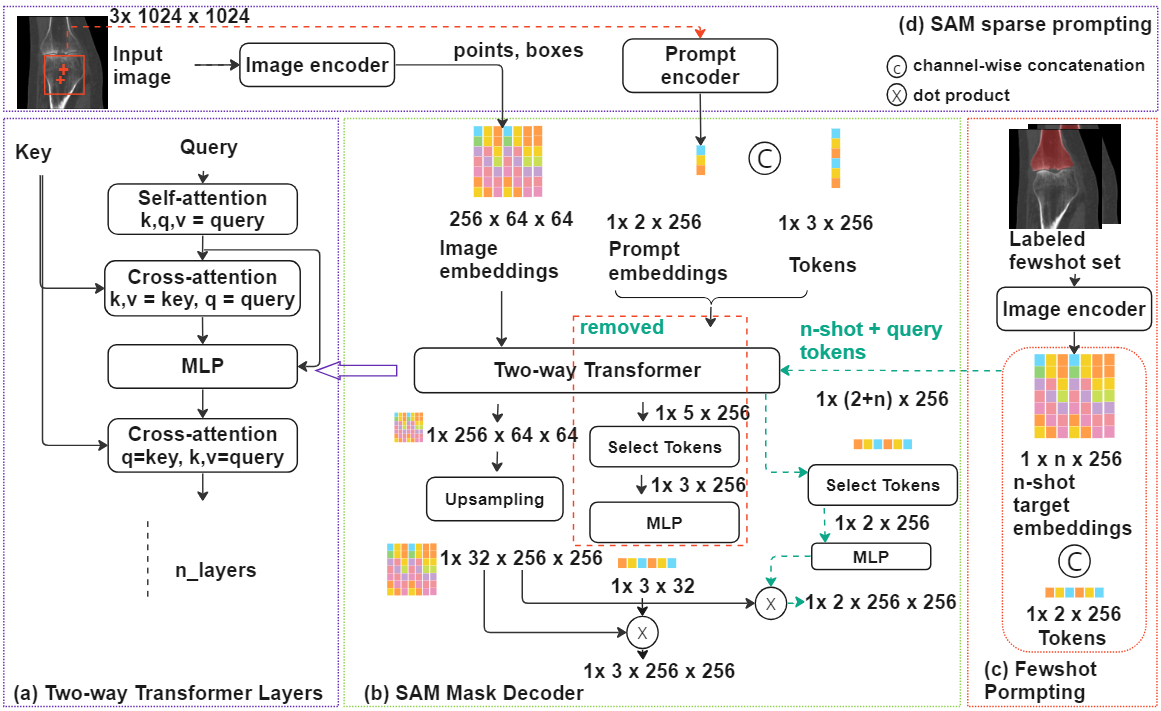}
   \caption{Mask decoder (b) in SAM using user prompts (d) and the proposed prompting method based on fewshot target embeddings derived from a set of labeled images (c). Two-way transformer layers allow both image embeddings and prompt embeddings to attend to each other's information (a). To maintain clarity, we have omitted the process of dense mask embeddings and positional encodings. Green dash lines indicate the process in the mask decoder that are modified for fewshot finetuning. And the part in the box marked by the red dash line is removed in the modified mask decoder as the proposed method does not use user prompt embeddings.}
   \label{fig:fewshot_overview}
\end{figure*}

SAM takes a 2D image with dimensions of 1024 $\times$ 1024 and RGB channels as input. The first step of SAM is to utilize its image encoder, a vision transformer \cite{dosovitskiy2020image,he2022masked}, to extract image embeddings from the input image. The resulting image embeddings are obtained at a down-sampled resolution of 64 $\times$ 64. SAM incorporates user input prompts, including points, bounding boxes, and masks, and encodes objects and their positional information into prompt embeddings to identify and locate objects within the image. These prompt embeddings serve as queries for the mask decoder, which is based on MaskFormers \cite{cheng2021per,cheng2022masked}. The mask decoder (illustrated in Figure~\ref{fig:fewshot_overview}) employs attention mechanisms to capture the correlations between the queries (prompt embeddings with tokens) and keys (image embeddings with encoded positional information). This enables the retrieval of relevant information stored in the image embeddings (values). The mask decoder comprises multiple layers of two-way transformers, as depicted in Figure~\ref{fig:fewshot_overview} (a). These transformers incorporate self-attention and cross-attention layers, allowing both the image and prompt embeddings to attend to each other's information. 

\subsection{Prompting with SAM}
SAM offers two modes: automatic mode and prompting mode. In automatic mode, users do not need to provide any input. The algorithm generates a grid of uniformly distributed points on the input image, which serve as prompts for segmentation. The auto-segmentation mode is not suitable for anatomical segmentation tasks as it lacks alignment with anatomical entities and segments anything in the image.

A more targeted approach is to use the prompting mode, which allows users to interact with the algorithm by providing various types of prompts such as points, bounding boxes, and masks to indicate the location of the target objects (skipping the text-based prompting because we do not find it in the open source release \footnote{https://github.com/facebookresearch/segment-anything}. In point prompting, users can provide multiple points to indicate the foreground and background areas. An alternative is to provide bounding boxes as prompts. Users can specify the coordinates of the top-left and bottom-right corners of the bounding boxes to indicate the regions of interest. This approach has shown to yield improved results when adapting SAM for medical image segmentation over using points prompting \cite{ma2023segment}.

\subsubsection{Fewshot Fine-tuning for SAM Adaption}
We propose a fewshot fine-tuning method for adapting SAM to segment anatomical structures from medical images. Instead of relying on user-provided prompts, our method utilizes SAM's image encoder to extract target embeddings from a set of fewshot images that are labeled for the specific segmentation task.

Given a fewshot set $D_{L} ={(x_{i},y_{i})}^{N_{L}}{i=1}$, where ${N_{L}}$ is the number of labeled images in $D_{L}$, $x_{i}$ denotes $i$-th image, and $y_{i}$ denotes the corresponding segmentation ground truth. Both $x_{i}$ and $y_{i}$ are 2D images ($x_{i}, y_{i} \in R^{W\times H}$) with spatial size $W \times H$. We first run SAM's image encoder on each image to obtain image embeddings $z_{i} \in R^{256 \times W^{'} \times H^{'}}$ at a 16$\times$ downsampled resolution ($W^{'} = W / 16, H^{'} = H / 16$).

To align the resolution of the embeddings with the segmentation ground truth, we downsample the corresponding ground truth $y_{i}$ to $\hat{y}_{i}, \hat{y}_{i} \in R^{W^{'} \times H^{'}}$. For each anatomical label $l$, we compute the target embedding $\hat{z}_{i}^{l} \in R^{256}$ by averaging the embedding vectors only within the downsampled mask corresponding to label $l$, applying the formula $\hat{z}_{i}^{l} =\sum [(\hat{y}_{i} = l) * z_{i}] / \sum (\hat{y}_{i} = l)$, where the summation iterative across all spatial locations in $\hat{y}_{i}$, $(\hat{y}_{i} = l)$ is the binary ground truth for label $l$, and $*$ denotes element-wise multiplication. Finally, all fewshot target embeddings for the set $D_{L}$ are in $R^{N_{L}\times C\times 256}$, where $C$ are the number of labels.

Fewshot target embeddings are concatenated with query tokens as one part of the input to the modified mask decoder; the other input is the image embeddings (refer to Figure~\ref{fig:fewshot_overview} (a)).
We modify the mask decoder to accept fewshot target embeddings (as indicated by the green dashed lines in Figure~\ref{fig:fewshot_overview} (b)) rather than user-defined prompts (as shown by the red box with a dashed line border in Figure~\ref{fig:fewshot_overview} (a)). Practically, the only change is the number of output tokens to fetch from the two-way transformer layers (two versus three in the original SAM's mask decoder), which is also the number of input tokens to the MLP. 
The use of query tokens has the same intuition as that in the original transformer \cite{vaswani2017attention}. In our case, for each label, we query two masks (foreground and background) using two tokens to represent a one versus the rest pixel-wise classification schema. At each layer, both the fewshot target embeddings and the image embeddings attend to each other using a cross-attention mechanism. This allows the model to capture the correlations between the fewshot target embeddings and the image embeddings in the embedding space. After the first layer, the fewshot embeddings with tokens are also self-attended. This self-attention mechanism helps refine the representation of the fewshot embeddings and incorporate relevant information across embeddings in the fewshot set. 

Same as the original SAM's mask decoder, the image embeddings enriched from the two-way transformer layers are upsampled and reshaped to a lower dimension. This reduces the dimensions of the image embeddings from $R^{256}$ to $R^{32}$, enabling more efficient computation for resolution reconstruction. Accordingly, the fewshot target embeddings and tokens are reshaped into the same reduced dimension ($R^{32}$), allowing for dot-product operations with the upsampled image embeddings. The final segmentation prediction for each anatomical label from the modified mask decoder has two channels (foreground and background) after spatially resizing into the original resolution (1024 $\times$ 1024).  

\subsubsection{Rationale Behind Fewshot Prompting}
In SAM, the mask decoder design, similar to the transformer decoder in MaskFormer \cite{cheng2021per}, aims to retrieve relevant information from the image embeddings (Values) using keys that are partially represented by positional encodings. The process involves generating queries based on the positional encodings derived from user-provided location information, such as points or bounding boxes. These positional encodings (queries) are then matched with the positional encodings (keys) stored within the image embeddings. By doing so, the regions within the image embeddings that correspond to the user input can be fetched and transformed into segmentation predictions. This mechanism allows SAM to effectively utilize the spatial relationships between the user-provided location information and the image embeddings for accurate segmentation.

In our proposed method, we eliminate the need for a prompt encoder and instead utilize fewshot target embeddings for segmentation. The labels of the target embeddings are propagated to the embeddings extracted from a test image. Rather than explicitly designing a nearest neighbor matching label propagation schema, our method leverages the power of the two-way transformer layers within SAM's mask decoder to facilitate optimal matches between the fewshot target embeddings and the image embeddings extracted from a test image. The target embeddings act as queries, allowing the retrieval and transformation of information stored in the image embeddings generated by SAM's image encoder into segmentation predictions.
In essence, the cross-attention maps between target embeddings and image embeddings in the two-way transformer layers represent the pair-wise similarity between queries and keys in the embedding space, while the distance metrics and mapping techniques to measure these similarities were learned via the fine-tuning process.

\section{Experiments}
\begin{table*}[thbp]
  \centering
  \begin{tabular}{|c|c|c|c|c|c|c|c|c|}
    \hline
 \multirow{2}{*}{Dataset} & \multirow{2}{*}{\makecell{Anatomy \\ (Modality)}} & \multicolumn{3}{c|}{\#Subjects} &  \multirow{2}{*}{\makecell{Spacings \\ (mm)}} & \multirow{2}{*}{Axes} & \multirow{2}{*}{Intensity clip} & \multirow{2}{*}{\makecell{2D slices \\in training}}\\\cline{3-5}
    &&\#total &\#train &\#test&&& & \\\hline
    \multirow{2}{*}{Internal} & Tibia (CT)& \multirow{2}{*}{3308} & \multirow{2}{*}{3008} & \multirow{2}{*}{300} &  \multirow{2}{*}{isotropic 1.0} &\multirow{2}{*}{coronal} & \multirow{2}{*}{[-500, 1300]} & \multirow{2}{*}{3008}\\\cline{2-2}
     & Femur (CT)&&&  & & & & \\\hline
    \multirow{2}{*}{AMOS22 \cite{ji2022amos}} & Aorta (CT) & \multirow{2}{*}{360} & \multirow{2}{*}{288} & \multirow{2}{*}{72} &  \multirow{2}{*}{in-plane 0.8} & \multirow{2}{*}{axial} & \multirow{2}{*}{[-160, 240]} & \multirow{2}{*}{3561} \\\cline{2-2}
     & Postcava (CT)&&&& & & &\\\hline
    MSD \cite{antonelli2022medical}&Left atrium (MRI)&20&15&5 & isotropic 1.0 & axial & [0.5\%, 99.5\%] & 583 \\\hline
    Verse20 \cite{sekuboyina2021verse} & Vertebrae (CT) & 60 & 47&13 & in-plane 1.0 & coronal &[-500, 1300] & 691\\\hline
  \end{tabular}
  \caption{Data collection and processing parameters across six different anatomies on four different data sources. We listed each set's total number of subjects and splits, including the number of 2D slices used in training. After resampling 3D images into fixed isotropic or in-plane resolutions, the 2D slices were extracted on the sampling axes. Clipping ranges are given for rescaling CT intensity values. For MRI scans, percentiles are used to exclude outliers.}
  \label{tab:data}
\end{table*}
\subsection{Datasets}
We collected three publicly-available datasets for anatomical segmentation on CT or MRI images. They are AMOS22 \cite{ji2022amos}, MSD \cite{antonelli2022medical}, and Verse20 \cite{sekuboyina2021verse}. We also collected one large-scale CT dataset internally. In total, all dataset contains 3,748 subjects, covering six anatomical structures, consisting of the tibia, femur, vertebrae, heart, aorta, and postcava. Datasets are summarized in Table~\ref{tab:data}. Because SAM operates on 2D images, we sampled 2D slices from 3D images for training and testing. The slices were selected on the predefined axes after resampling scans into either fixed isotropic spacings or in a fixed in-plane resolution while keeping the original $z$-spacing (Table~\ref{tab:data}). For simplicity, we avoid the instance-segmentation problem in the Verse20 dataset, e.g., all vertebrae bodies in the Verse20 dataset are considered one anatomical label.   

\subsection{Methods Comparison}
Three methods are compared: SAM (using point or box prompts), nnU-Net \cite{isensee2018nnu}, and the proposed fewshot fine-tuning methods. We extract points and bounding boxes from the segmentation ground truth to mimic the user providing accurate points or boxes for prompting SAM. Note that this represents an idealized setting, where ground truth is leveraged to generate comprehensive prompts. For each anatomical label, we first perform connected component analysis. Then for each connected component, we compute a single point within the component mask, using the coordinate where the value in the distance map function is the highest. The distance map function ($DMF_{x}$) for a given location $x$ to the segmentation ground truth $\Omega$ is computed as the follows: 
\begin{equation}\label{eq:sdf}
  DMF_{x} =
    \begin{cases}
      \inf \lVert x - d\Omega \rVert_{2} & \text{if } x \in \Omega \\
      0 & \text{others}
    \end{cases}       
\end{equation}
where $\lVert x - d\Omega \lVert_{2}$ is the euclidean distances between voxels coordinates $x$ and coordinates in the ground truth object boundary set $d\Omega$. The bounding box for each component is simply derived from the segmentation ground truth as the top left and bottom right corners. Given an anatomy in 2D cross-section view with $c$ connected component, the final point prompts are a sequence of coordinates $c \times 2$ combined with $c$-dimensional binary point labels (one if present in foreground, otherwise zero). The box prompts are $c \times 2 \times 2$. For the Verse20 dataset, we have, on average, six connected components (vertebrae instances) on each 2D image. In such cases, manual prompting may be infeasible for practical usage. To showcase SAM's zeroshot capability, SAM (using point or box prompts) is only executed on images on the test set without retraining or fine-tuning.  

nnU-Net was trained using all 2D slices from the training split. We tuned training hyper-parameters for each task separately using a validation set with 10\% of images within the training set. 

In our proposed fewshot fine-tuning method, we perform fine-tuning on the modified mask decoder using fewshot subsets created with varying sizes, specifically 5, 20, and 50 examples from the training set. During fine-tuning, the modified mask decoder is trained only using the fewshot subset, while the fully-supervised nnU-Net is trained on all available training examples. The images in a fewshot subset are used to extract fewshot target embeddings using SAM's image encoder. Once extracted, these fewshot target embeddings can be cached as part of model storage, allowing them to be used as prompts during test time. This caching mechanism also makes subsequent iterations of fine-tuning efficient. 

\subsection{Metrics}
We report the intersection over union (IoU) and average symmetric surface distance (ASSD) as the metrics for evaluating the segmentation performance. IoU metrics were computed using the resampled spacings as shown in Table~\ref{tab:data}. ASSD metrics were reported in milli-meters. 

\subsection{Experiment Details}
The model parameters were optimized using the Adam optimizer \cite{kingma2014adam} with an initial learning rate of $10^{-4}$ and decay rates $\beta_1$ set to 0.9 and $\beta_2$ set to 0.99. The network parameters were initialized using He initialization \cite{he2015delving}. The nnU-Net method was trained for a maximum of 200 epochs for all tasks, with training stopping when the metrics on the validation set did not show improvement for ten epochs. Data augmentations were randomly applied during training, including intensity scaling and shifting, contrast stretching, Gaussian additive noise, spatial flipping, and resizing after cropping. For fine-tuning the modified mask decoder, a maximum of 50, 80, and 100 iterations were performed when the fewshot set consisted of 5, 20, and 50 labeled images, respectively. The images in the fewshot set were selected to best represent the target anatomy's appearance. All experiments were conducted on a machine with 4 NVIDIA Tesla K80 GPUs, 24 CPU cores, and 224 GB RAM. The fine-tuning experiments were completed within three hours using the cache mechanism. 

\begin{table*}[htbp]
\centering
\begin{tabular}{|c|c|c|c|c|c|c|c|}
 \hline
\multirow{2}{*}{Anatomy} & \multirow{2}{*}{Metrics} & \multicolumn{2}{|c|}{SAM} & \multicolumn{3}{|c|}{Ours (n-shot)} & \multirow{2}{*}{nnUNet} \\\cline{3-7}
&& Point & box &n=5 &n=20 & n=50&\\\hline
\multirow{2}{*}{Femur} &IoU \% & $66.4\pm26.2$ & $97.2\pm1.3$ & $96.8\pm2.2$ & $97.6\pm3.3$ & $97.8\pm2.0$ & \textbf{97.8$\pm$ 0.9}\\\cline{2-8}
& ASSD & $10.4\pm9.0$ & $1.3\pm0.7$ & $1.5\pm1.2$ & $1.1\pm1.6$ & $1.00\pm1.1$ & \textbf{0.6$\pm$0.7} \\\hline
\multirow{2}{*}{Tibia} & IoU \% & $50.3\pm17.9$ &$96.3\pm2.3$& $95.3\pm5.7$ & $96.8\pm4.6$ & $97.1\pm3.4$ & \textbf{97.5$\pm$2.6}  \\\cline{2-8}
& ASSD& $17.5\pm10.3$ & $1.48\pm1.1$ & $2.0\pm2.9$& $1.2\pm2.0$ & $1.1\pm1.5$ & \textbf{0.7$\pm$1.2}\\\hline
\multirow{2}{*}{Left Atrium} &IoU \%  & $66.4 \pm 26.0$ & $88.4\pm6.5$ & $82.4\pm13.2$ & $83.1\pm12.2$ & \textbf{89.1$\pm$8.7} &$88.9\pm7.9$ \\\cline{2-8}
& ASSD& $10.4 \pm 7.8$ & $2.9\pm1.4$ &$6.56\pm6.6$ & $5.53\pm5.4$ &$2.12\pm2.0$& \textbf{0.6$\pm$0.6} \\\hline
\multirow{2}{*}{Aorta} & IoU \%  & $53.3\pm31.2$& $88.8\pm7.6$ &$82.4\pm8.9$& $90.1\pm8.4$&$90.8\pm8.3$ & \textbf{94.2$\pm$5.7} \\\cline{2-8}
& ASSD & $29.1\pm27.2$& $0.8\pm0.6$ &$7.4\pm4.7$ &$1.8\pm1.6$ & $1.0\pm1.4$ & \textbf{0.4$\pm$0.8} \\\hline
\multirow{2}{*}{Postcava} & IoU \%  & $69.4\pm17.3$ & \textbf{79.2$\pm$8.9} &$73.3\pm11.2$ & $76.1\pm 10.7$&$77.7\pm9.1$& $78.1 \pm 6.1$  \\\cline{2-8}
& ASSD & $12.8\pm9.4$& \textbf{2.1$\pm$1.6} &$8.1\pm5.3$ &$7.8\pm3.6$ & $5.2\pm2.4$ & $2.7\pm2.5$ \\\hline
\multirow{2}{*}{Vertebrae} & IoU \%  & $65.7\pm34.1$ &\textbf{93.4$\pm$2.8} &$85.1\pm6.9$&$89.3\pm10.7$& $92.9\pm5.4$ & $92.7\pm3.7$ \\\cline{2-8}
& ASSD& $29.9\pm40.4$ &\textbf{0.8$\pm$0.4} &$6.2\pm5.7$ & $4.5\pm4.9$& $3.2\pm3.5$ & $0.7\pm0.8$ \\\hline
\end{tabular}
\caption{Quantitative results of anatomical segmentation methods on various anatomies. Three set of methods in comparison: SAM (point or boxes as prompts), the proposed few-shot methods trained with 5, 20, and 50 labeled images, and the fully-supervised nnUNet trained with full data. Best results are in bold. ASSD: average symmetric surface distance, IoU: intersection over union.}
\label{tab:mainresults}
\end{table*}

\subsection{Results}
\begin{figure*}[ht!]
  \centering
  \setlength{\tabcolsep}{0.001\textwidth}
  \begin{tabular}{cccccc}
   \includegraphics[width=0.165\linewidth, height=3.1cm]{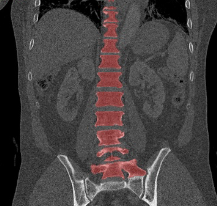}&
   \includegraphics[width=0.165\linewidth, height=3.1cm]{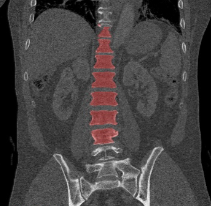}&
   \includegraphics[width=0.165\linewidth, height=3.1cm]{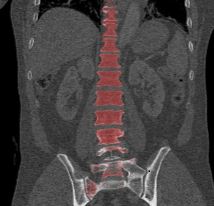}&
   \includegraphics[width=0.165\linewidth, height=3.1cm]{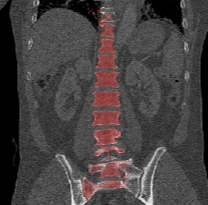}&
   \includegraphics[width=0.165\linewidth, height=3.1cm]{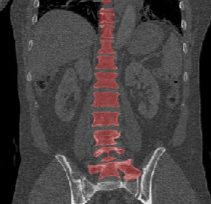}&
   \includegraphics[width=0.165\linewidth, height=3.1cm]{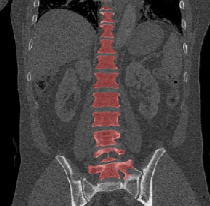} \\
   \includegraphics[width=0.165\linewidth, height=3.1cm]{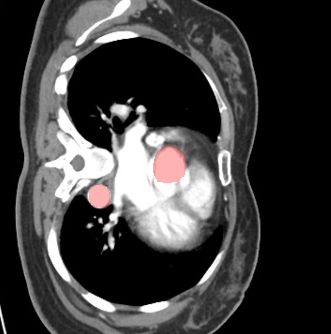}&
   \includegraphics[width=0.165\linewidth, height=3.1cm]{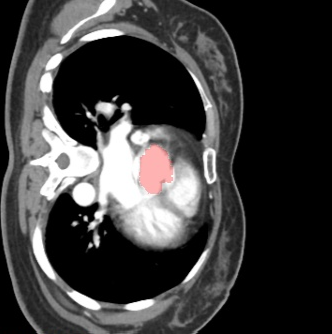}&
   \includegraphics[width=0.165\linewidth, height=3.1cm]{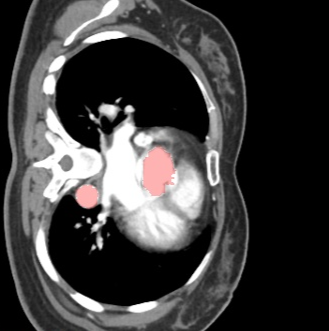}&
   \includegraphics[width=0.165\linewidth, height=3.1cm]{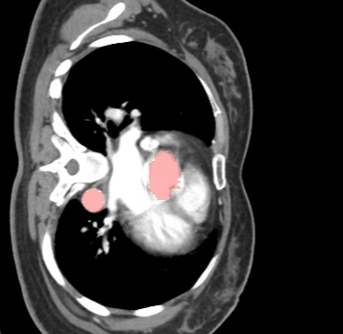}&
   \includegraphics[width=0.165\linewidth, height=3.1cm]{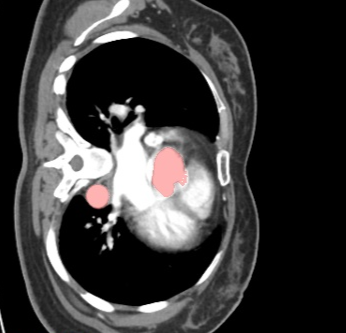}&
   \includegraphics[width=0.165\linewidth, height=3.1cm]{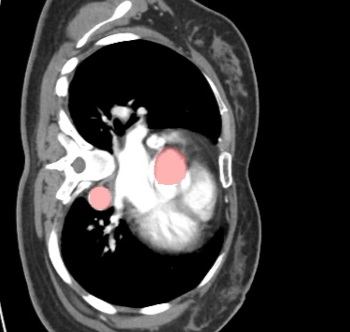} \\
   \includegraphics[width=0.165\linewidth, height=3.1cm]{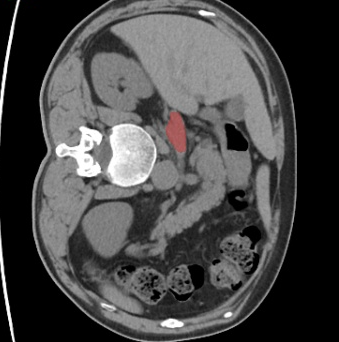}&
   \includegraphics[width=0.165\linewidth, height=3.1cm]{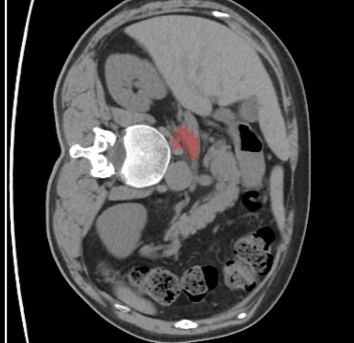}&
   \includegraphics[width=0.165\linewidth, height=3.1cm]{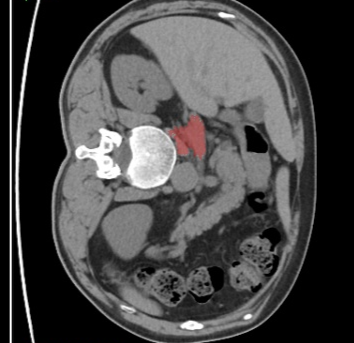}&
   \includegraphics[width=0.165\linewidth, height=3.1cm]{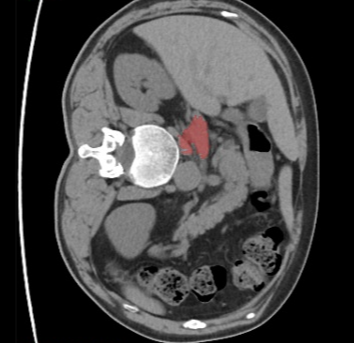}&
   \includegraphics[width=0.165\linewidth, height=3.1cm]{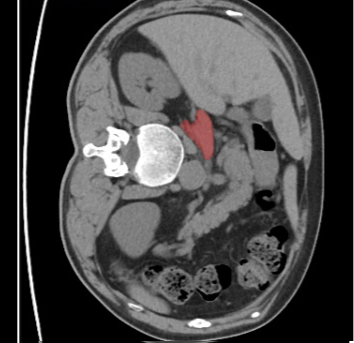}&
   \includegraphics[width=0.165\linewidth, height=3.1cm]{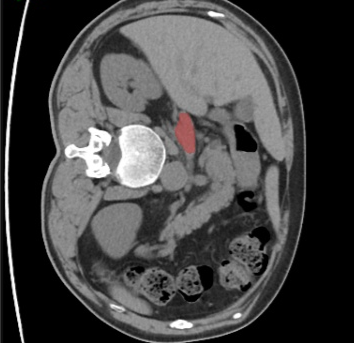} \\
   \includegraphics[width=0.165\linewidth, height=3.1cm]{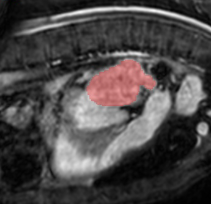}&
   \includegraphics[width=0.165\linewidth, height=3.1cm]{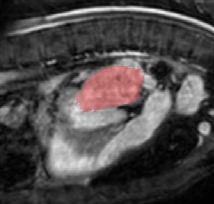}&
   \includegraphics[width=0.165\linewidth, height=3.1cm]{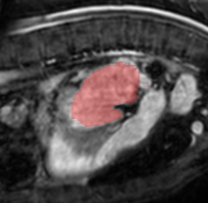}&
   \includegraphics[width=0.165\linewidth, height=3.1cm]{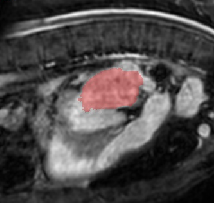}&
   \includegraphics[width=0.165\linewidth, height=3.1cm]{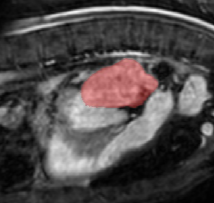}&
   \includegraphics[width=0.165\linewidth, height=3.1cm]{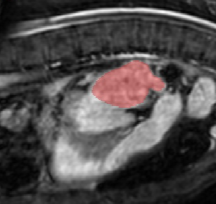} \\
   \includegraphics[width=0.165\linewidth, height=3.1cm]{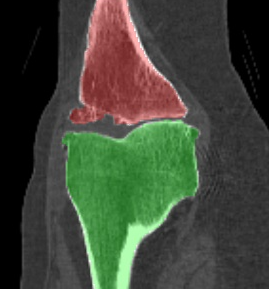}&
   \includegraphics[width=0.165\linewidth, height=3.1cm]{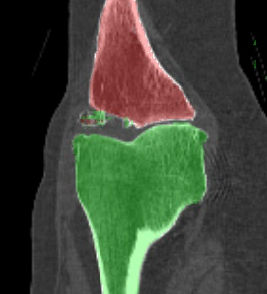}&
   \includegraphics[width=0.165\linewidth, height=3.1cm]{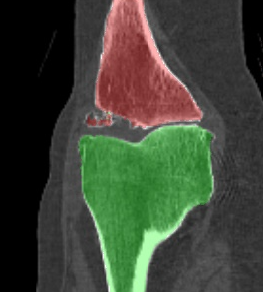}&
   \includegraphics[width=0.165\linewidth, height=3.1cm]{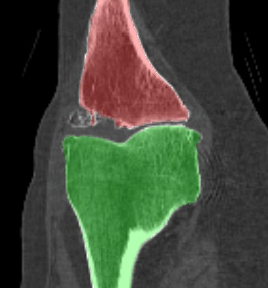}&
   \includegraphics[width=0.165\linewidth, height=3.1cm]{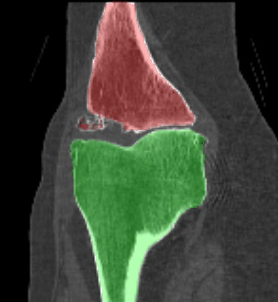}&
   \includegraphics[width=0.165\linewidth, height=3.1cm]{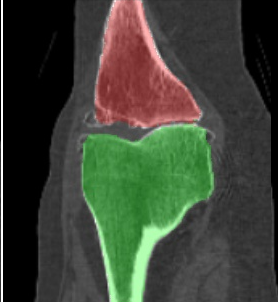} \\
\end{tabular}
   \caption{Qualitative results of the segmentation methods in comparison. The first column represents the ground truth segmentations, where all anatomies are visualized in red, except for the femur, which is visualized in green to differentiate it from the tibia. The subsequent columns depict the results of different segmentation methods: SAM with bounding box prompt (5th column), the fully supervised nnU-Net (6th column), and the proposed fewshot finetuning with 5, 20, and 50 labeled images (2nd-4th columns). Each row corresponds to a different case.}
   \label{fig:qualitative}
\end{figure*}

Table~\ref{tab:mainresults} shows that SAM with only point prompts exhibits suboptimal performance, with Intersection over Union (IoU) below 70\% and Average Symmetric Surface Distance (ASSD) exceeding 10 mm on all tasks. This is primarily due to the inherent ambiguity in the predictions, as demonstrated in Figure~\ref{fig:SAM_point_fail}. SAM utilizing bounding boxes as prompts demonstrates remarkable zero-shot capabilities, achieving the highest scores in the postcava and vertebrae among all methods. Although SAM with accurate bounding box prompts shows solid performance, qualitative results (Figure~\ref{fig:qualitative} 5th column, 5th row) show that SAM image embeddings, in general, do not suffice to represent pathological changes that do not belonging to the common anatomy. In the cases of knee joint segmentation, osteophytes (shown on the tibia, colored in red) are partially missed (marked in a yellow circle). As we do not update SAM's image encoder, this segmentation error shows that the image embeddings extracted by the pre-trained image encoder fail to describe osteophytes. Segmenting pathological changes itself is a challenging task as these changes may be under-represented in the data collection, given we also saw same under-segmentation in nnU-Net's results (Figure~\ref{fig:qualitative} 6th column, 5th row). Interestingly, our fine-tuning approach produces even better results than SAM with box prompts on segmenting the femur, tibia, left atrium, and aorta when using 50 labeled images, showing the proposed method's ability to segment difficult anatomical structures when sufficient images are labeled for fine-tuning. Crucially, the required number of labeled images are an order of magnitude less than to achieve similar results with the fully supervised nnU-Net approach, especially for knee segmentation. 

In contrast, the fine-tuning method using only 5 labeled images on Verse20 produces sub-optimal results, with IoU at 85.1 and ASSD at 6.2mm compared with box-prompting SAM at IoU 93.4 and ASSD 0.8mm. The drop in performance for segmenting vertebrae with five-shot fine-tuning is attributed to missing coverage of all vertebrae in 2D coronal views. Most 2D views cover thoracic and lumbar vertebra, while cervical and sacrum spines are under-represented in the sampled 2D training slices. Therefore, including all vertebrae with only 5 2D coronal slices is challenging.
For segmenting tubular structures such as the left atrium, aorta, and postcava, the fewshot method with 5 images also faced the challenges given these structures may look very different in 2D cross-section views, and such variations cannot be sufficiently summarized with only 5 images. By adding more labeled images to the fewshot set, the segmentation performance can be improved substantially.
Qualitative results show that the segmentation errors were mostly over-segmentation of adjacent non-target anatomies (such as ilium bone segmented in Figure~\ref{fig:qualitative} 1st row, 4th column not part of sacral vertebrae). Overall, the segmentation errors in the proposed approach may be caused by the SAM's image encoder not capturing sufficient anatomical semantics, especially around the boundaries of these anatomical structures. Over-segmentations mostly show that these features failed to distinguish adjacent anatomies, especially when they appear similar (e.g., left atrium versus the surrounding tissues in other parts of the heart).
On the other hand, we also observe that SAM's image embeddings are powerful enough to identify unrelated regions that are far from the target anatomy, even when they are similar in appearance. In abdomen CT images, many anatomical structures have similar intensity distributions, such as between the aorta and postcava. Figure~\ref{fig:qualitative} shows no false positives of labeling the aorta as postcava or vice versa in the fewshot fine-tuning results. 

Finally, the nnU-Net trained with all labeled images achieved the best overall segmentation performance, showing that training task-specific model still produces better segmentation results if a sufficient number of labeled images is available. \add{When comparing the fewshot SAM method to other SOTA methods leveraging SAM, our approach achieves a better IoU score on aorta segmentation than MSA 1-point DICE score although trails behind MedSAM 1-point performance (DICE score). Although IoU and DICE scores are different metrics, they are strongly positively correlated and thus enable qualitative comparison of these methods.}

\section{Discussion \& conclusion}

This study introduces a method for adapting SAM to anatomical segmentation tasks in medical images (CT, MRI). Our approach eliminates user prompts and relies on a few labeled images for prompting. We compare this approach with prompt-based SAM (points and boxes) and fully supervised nnU-Net methods. Crucially, accurate ground truth segmentations are leveraged to generate the bounding box prompts for SAM, representing highly idealized prompts that are unlikely to be generated for every sample without significant labeling effort. \add{This level of prompting can be expensive and sometimes infeasible for medical anatomical structure segmentation, especially when dealing with large sparsely-distributed entities, such as airways and vessels.} 

Remarkably, our fine-tuning approach with 5 labeled images achieves comparable results to SAM with (idealized) bounding box prompts for femur and tibia segmentation, highlighting the potential to reduce the amount of labeling effort required and still maintain accurate segmentation of anatomical structures. 
The reduced requirement for labeled images while maintaining good performance is a key strength of our method. Additionally, the fine-tuning process is efficient by caching computed image embeddings, allowing for reuse in repeated runs. Experimental results demonstrate the effectiveness of our method in segmenting various challenging anatomies in CT or MRI images, even possible with only 5 labeled images for training. \add{Finally, the utility of our fewshot SAM goes beyond medical image segmentation, providing a generic framework for token-query-based object detection and classification tasks outside of the medical image and segmentation domains.}

{\small
\bibliographystyle{ieee_fullname}
\bibliography{main}
}

\end{document}


\title{Supplementary Material For WACV2024 \\ SAM Fewshot Finetuning for Anatomical Segmentation in Medical Images}

\author{
  Weiyi Xie, Nathalie Willems, Shubham Patil, Yang Li, and Mayank Kumar\\
  Stryker AI Research\\
  {\tt\small \{weiyi.xie,nathalie.willems,shubham.patil,yang.li1,mayank.kumar\}@stryker.com}
}
\maketitle

\section{\add{Embedding Analysis}}
\begin{figure*}[t]
  \centering
  \setlength{\tabcolsep}{0.001\textwidth}
  \begin{tabular}{ccc}
   \includegraphics[width=0.33\linewidth]{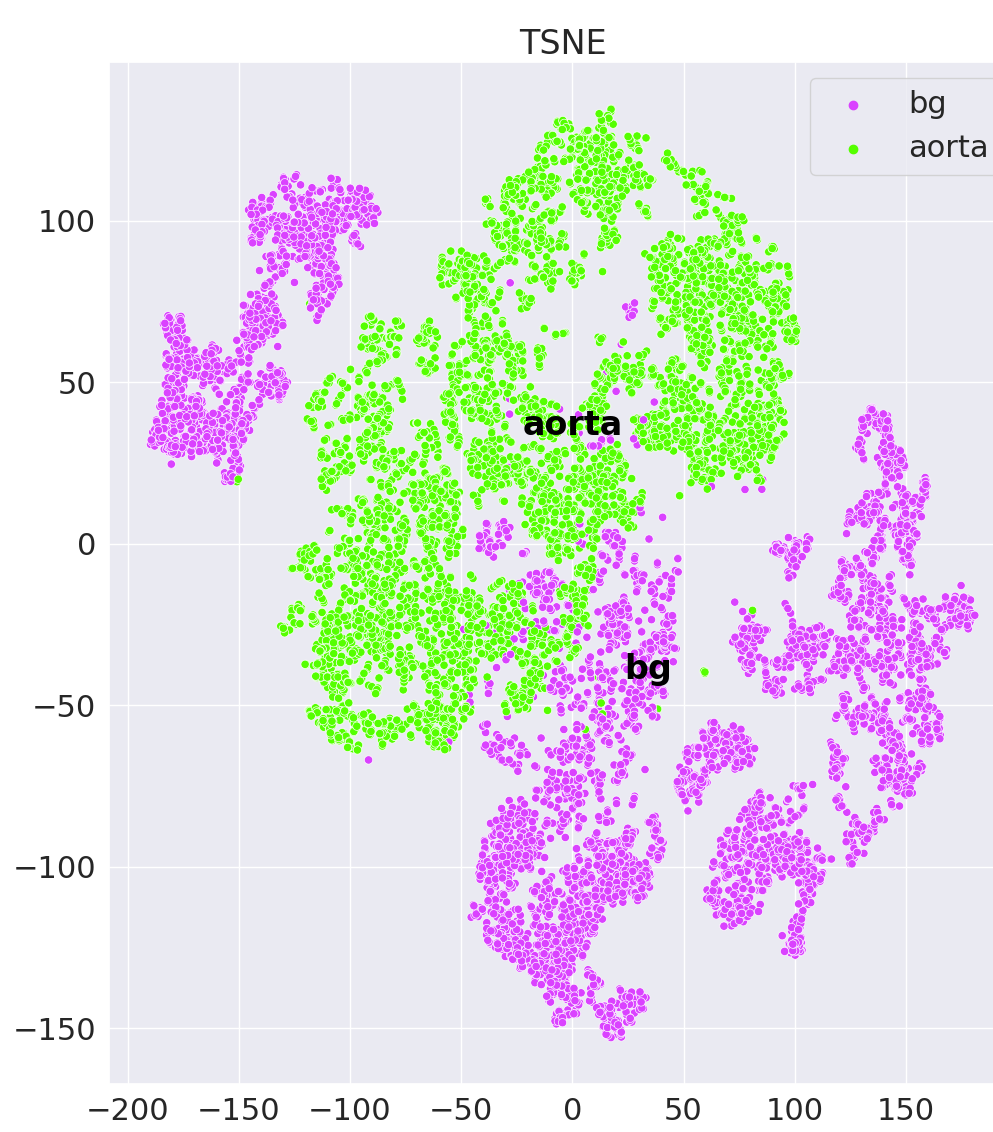}&
   \includegraphics[width=0.33\linewidth]{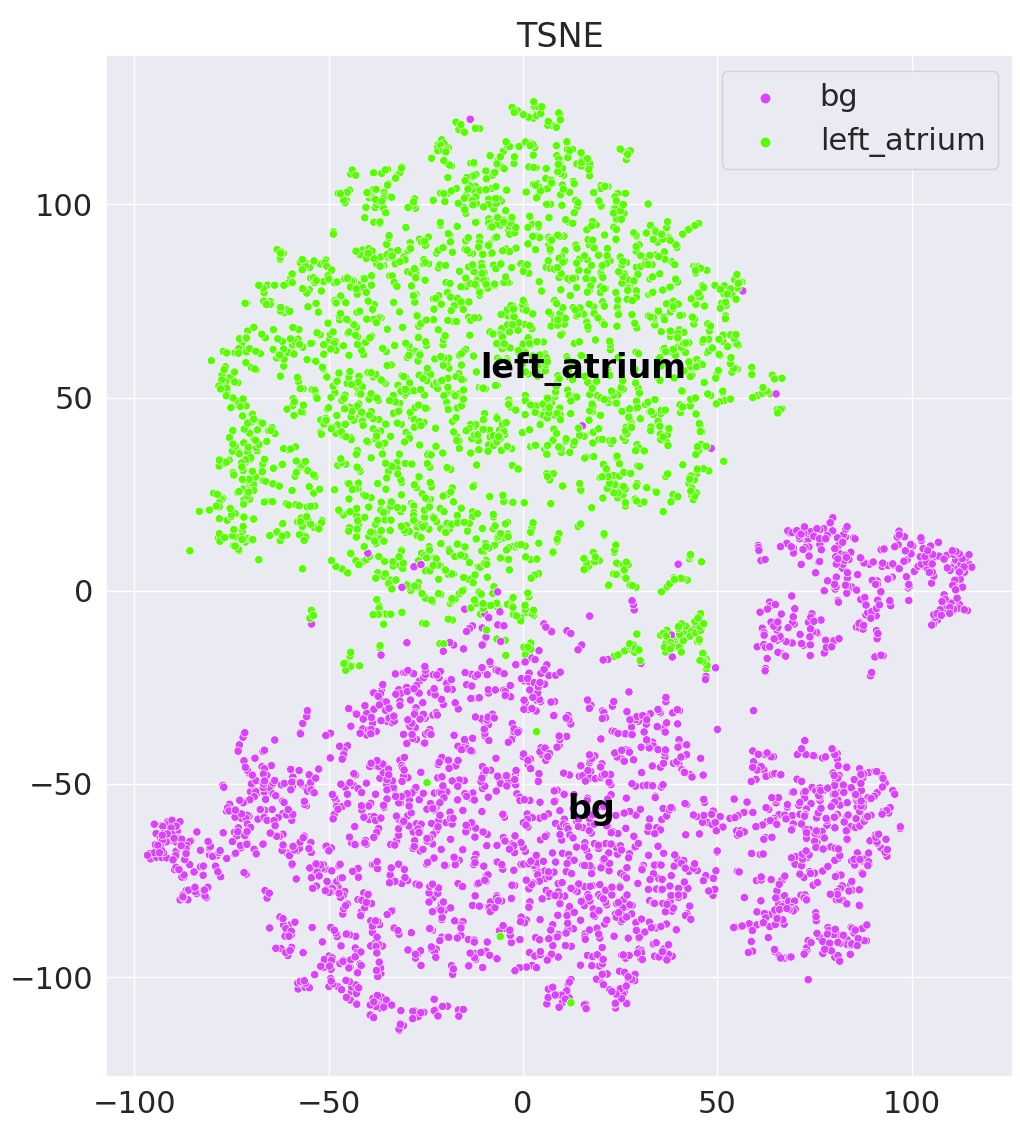}&
   \includegraphics[width=0.33\linewidth]{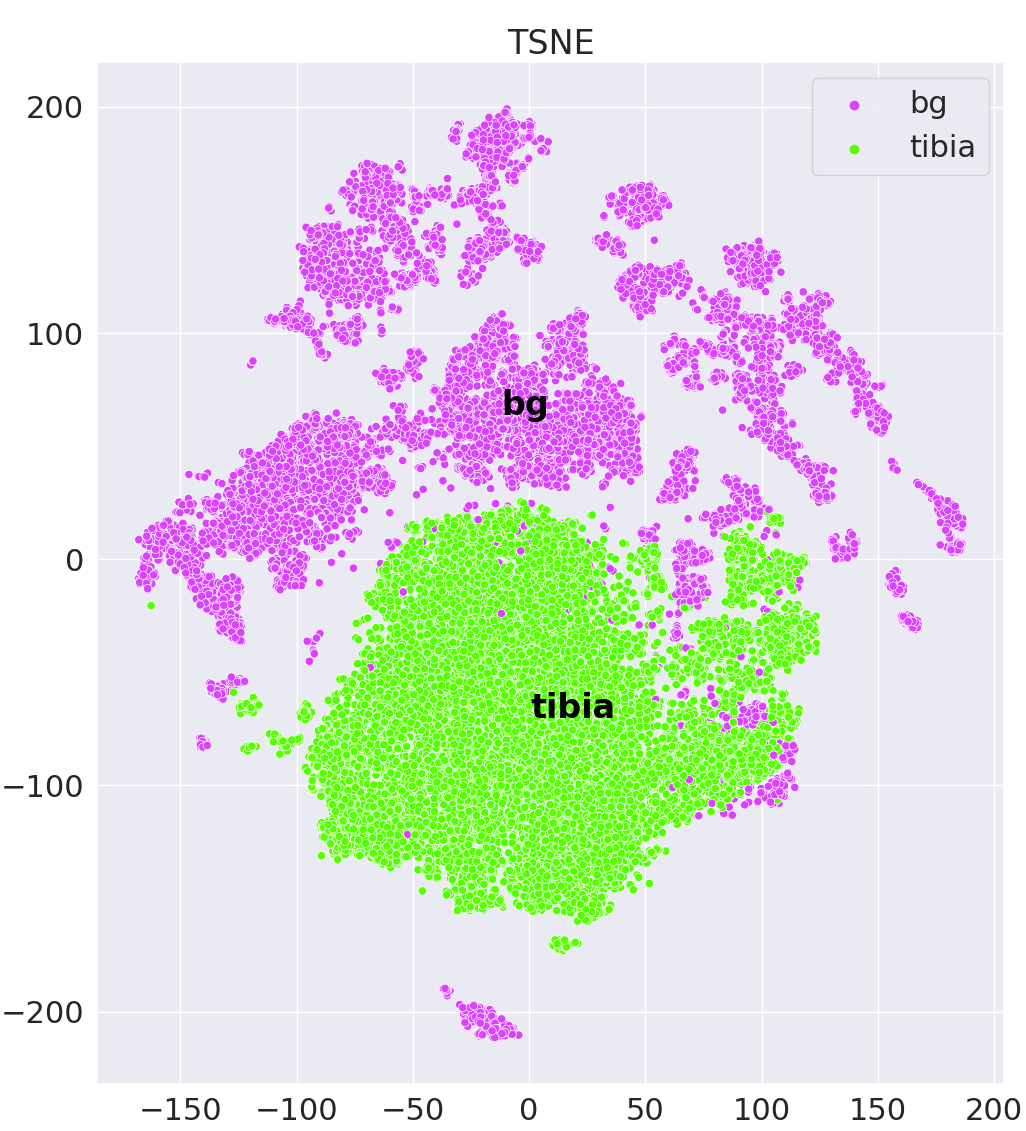}&\\
   \includegraphics[width=0.33\linewidth]{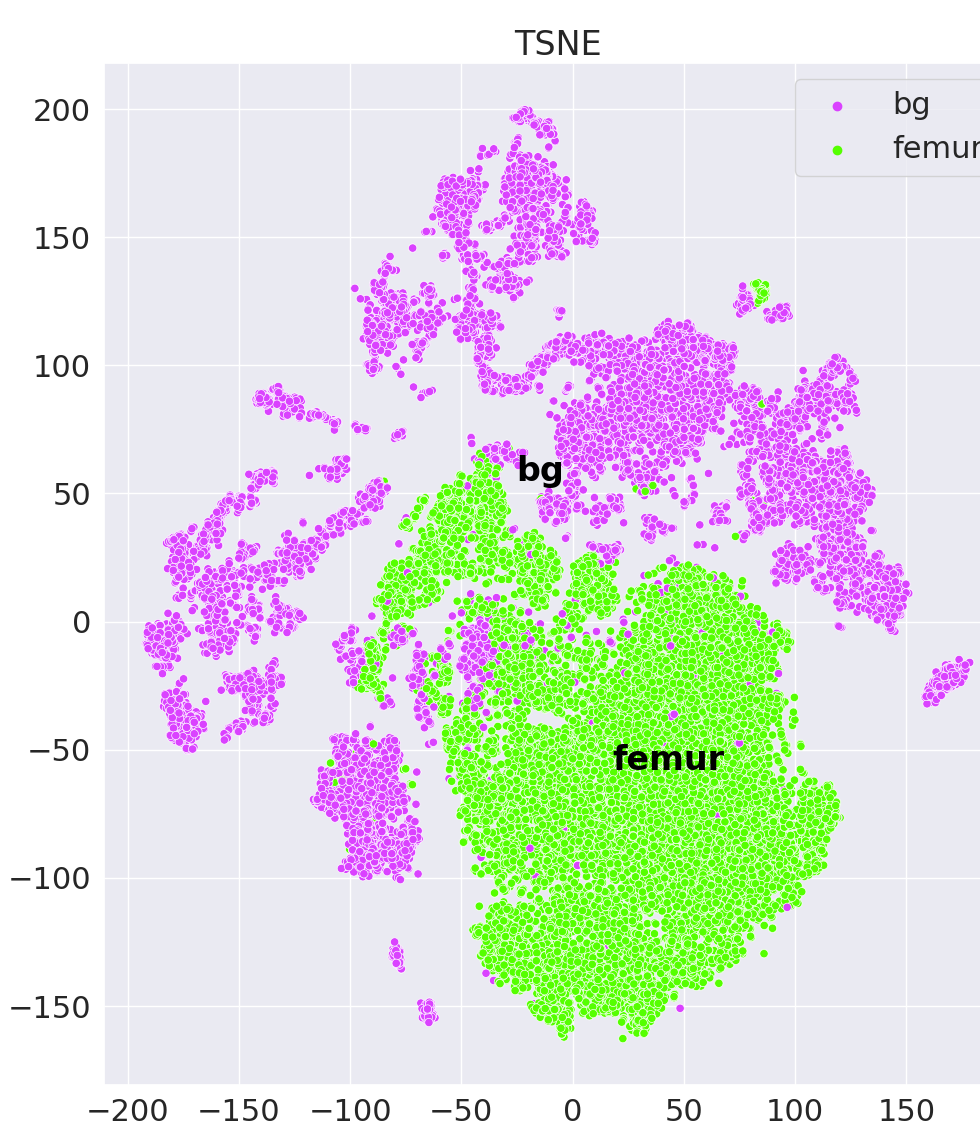}&
   \includegraphics[width=0.33\linewidth]{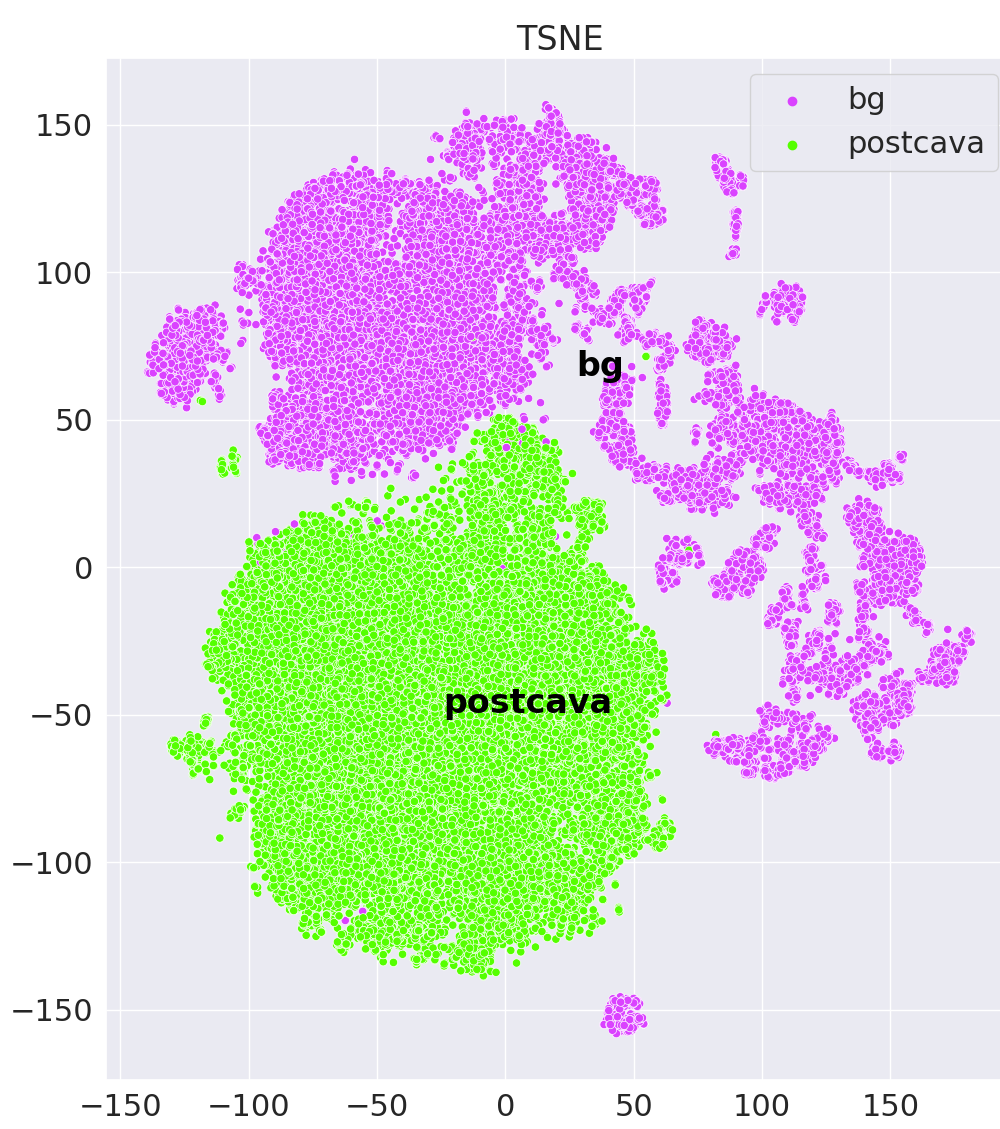}&
   \includegraphics[width=0.33\linewidth]{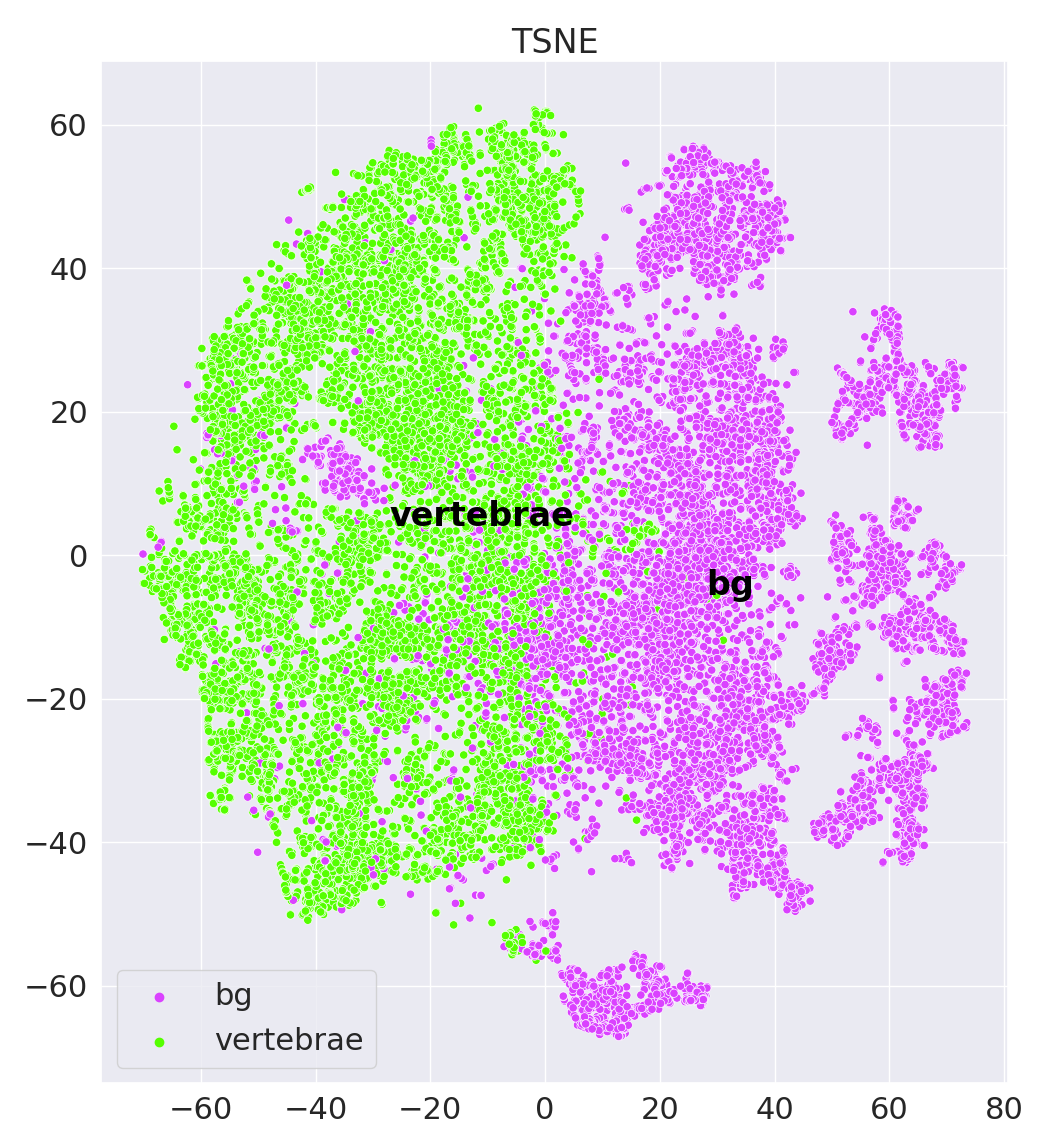}
\end{tabular}
   \caption{t-SNE plots showing the feature embeddings generated by the image encoder in the SAM for separating the anatomy against the background among tasks. The first row shows the class separation for aorta, left atrium and tibia from the left to the right, while the second row shows femur, postcava, and vertebrae class separation from the left to the right.}
   \label{fig:tsne}
\end{figure*}

\add{We present t-SNE plots as a means to visualize feature embeddings generated by the image encoder in SAM across all datasets utilized in this study. These t-SNE plots were created using a perplexity of 25 and 5000 iterations. We sampled equal number (n=50) of feature embeddings per class for each task on the test set images from SAM's image encoder.
As illustrated in Figure~\ref{fig:tsne}, the majority of classes exhibit clear separation in the embedding space. However, for the knee classes (tibia and femur), the separation within the embedding space is comparatively less distinct. This observation suggests the presence of ambiguities when utilizing SAM's feature embedding to distinguish between femur and tibia.
Nonetheless, it is worth noting that the proposed mask decoder and fine-tuning strategy appear to mitigate these ambiguities effectively. This is evidenced by the relatively high segmentation performance achieved for tibia and femur after fine-tuning, especially when compared to the results obtained through the fully-supervised method (refer to Table 2).}

\section{Number of Labeled Images in Fine-Tuning}
\begin{figure*}[t]
  \centering
  \setlength{\tabcolsep}{0.001\textwidth}
  \begin{tabular}{cc}
   \includegraphics[width=0.5\linewidth]{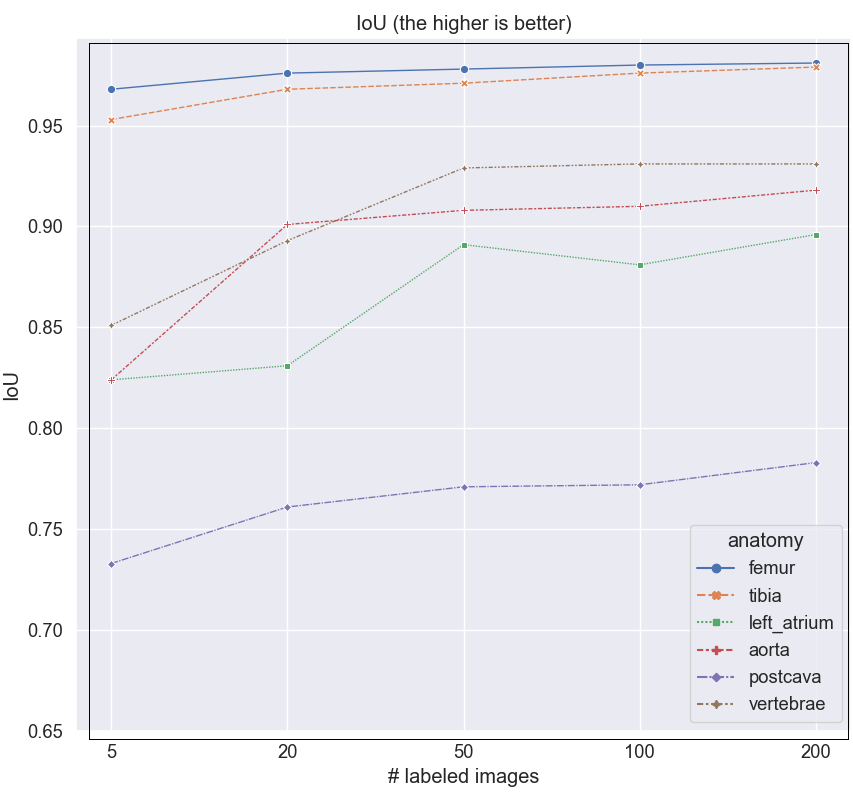}&
   \includegraphics[width=0.5\linewidth]{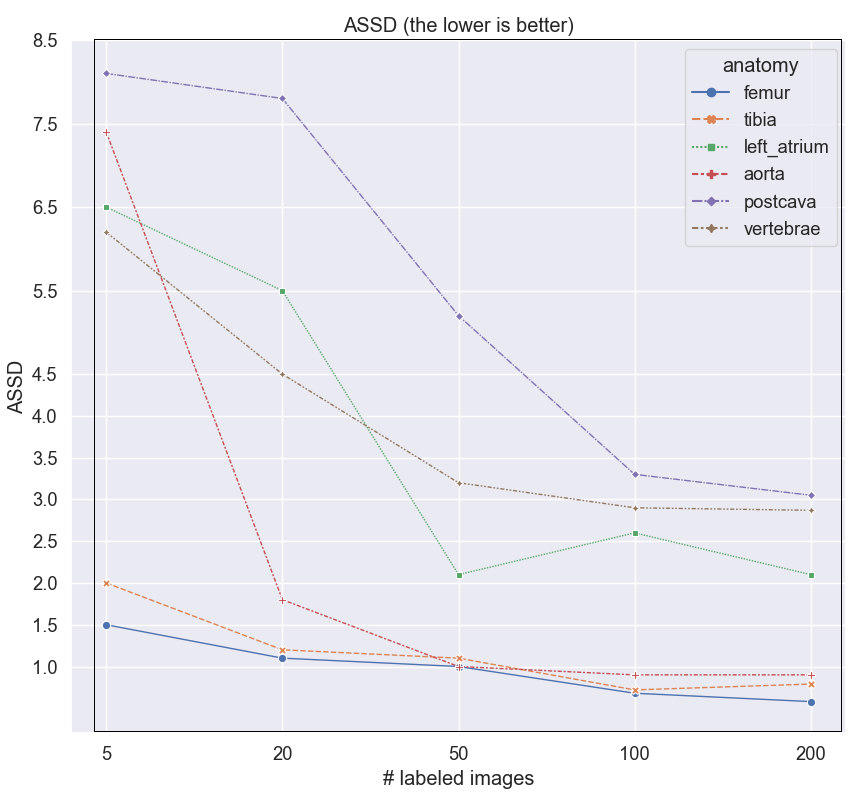}
\end{tabular}
   \caption{Incorporating a greater number of labeled images during the fine-tuning process generally enhances segmentation performance across various anatomical structures in this study. This improvement is manifested by an increase in Intersection over Union (IoU) on the left and a reduction in Average Surface Symmetric Distance (ASSD) on the right.}
   \label{fig:sampling}
\end{figure*}

In this section, we undertake a comparative analysis of the few-shot fine-tuning method, leveraging 5, 20, 50, 100, and 200 labeled images, in order to extend our investigation, as initially presented in Table~2. The results, depicted in Figure~\ref{fig:sampling}, clearly demonstrate that the inclusion of a greater number of labeled images generally enhances segmentation performance, as assessed through the IoU and ASSD metrics.
Notably, when considering femur and tibia, the IoU achieved with 200 labeled images stands at an impressive 98.1 \% and 97.9 \%, surpassing the performance of the fully-supervised method, which records 97.8 \% and 97.5 \%, respectively. For the segmentation of the left atrium, employing 200 labeled images yields an IoU of 89.6 \%, compared to the fully-supervised method's IoU of 88.9 \%. Similarly, when focusing on vertebrae segmentation, employing 200 labeled images results in an IoU of 93.1\%, surpassing the fully-supervised method's performance at 92.7\%.
It is worth noting, however, that this trend does not hold uniformly for all anatomical structures investigated in this study. For postcava and aorta, achieving fully-supervised performance may require adding more labeled images than 200 in the fine-tuning process.
